# Design Space Exploration of Neural Network Activation Function Circuits

Tao Yang, Yadong Wei, Zhijun Tu, Haolun Zeng, Michel A. Kinsy, *Member IEEE*,
Nanning Zheng, *fellow IEEE* and Pengju Ren, *Member IEEE*

*Abstract*—The widespread application of artificial neural networks has prompted researchers to experiment with FPGA and customized ASIC designs to speed up their computation. These implementation efforts have generally focused on weight multiplication and signal summation operations, and less on activation functions used in these applications. Yet, efficient hardware implementations of nonlinear activation functions like Exponential Linear Units (ELU), Scaled Exponential Linear Units (SELU), and Hyperbolic Tangent (tanh), are central to designing effective neural network accelerators, since these functions require lots of resources. In this paper, we explore efficient hardware implementations of activation functions using purely combinational circuits, with a focus on two widely used nonlinear activation functions, i.e., *SELU* and *tanh*. Our experiments demonstrate that neural networks are generally insensitive to the precision of the activation function. The results also prove that the proposed combinational circuit based approach is very efficient in terms of speed and area, with negligible accuracy loss on the MNIST, CIFAR-10 and I$_{\text{MAGE}}$N$_{\text{ET}}$ benchmarks. Synopsys Design Compiler synthesis results show that circuit designs for *tanh* and *SELU* can save between $\times$**3.13** $\sim$ $\times$7.69 and $\times$4.45 $\sim$ 8.45 area compared to the LUT/memory based implementations, and can operate at 5.14GHz and 4.52GHz using the 28nm SVT library, respectively. The implementation is available at: https://github.com/ThomasMrY/ActivationFunctionDemo.

## I. Introduction

Artificial neural networks (ANN) are deployed in a wide range of applications, such as image recognition, speech recognition, and natural language processing. Speeding up neural network inference and reducing power consumption have become essential in order to enable ANN adoption in edge devices where low-power and low-latency are required. Current CPUs and GPUs are ill-suited for this class of devices, leading many researchers to pursue custom FPGA or ASIC accelerators.

ANNs consist of neurons, which sum incoming signals and apply an activation function, and connections, which amplify or inhibit passing signals. When the neuron's activation function is nonlinear, the two-layer neural network becomes a universal function approximator [1]. Various nonlinear equations, such as sigmoid, logistic, tanh, Rectified linear unit (ReLU), Scaled Exponential Linear Unit (SELU), etc. [2] have been used to implement activation functions. Researchers in [3] show that nonlinear activation functions affect the learning and generalization capabilities of ANNs.

The rationale for focusing on the efficient implementation of exponential functions is twofold: (a) exponential functions are used in several activation functions, such as *ELU*, *SELU*, *tanh*, and *sigmoid*, and (b) the ELU [4] and SELU [5] functions have been shown (i) to significantly decrease training time, (ii) to push mean activations closer to zero, (iii) to not require batch normalization, and (iv) to alleviate the vanishing gradient problem. For example, the SELU activation function provides lower and upper bounds on the gradient variance and removes the vanishing/exploding gradient problem. Therefore, we expect a wider adoption of these activation functions in the future and attempts to reduce their hardware area, latency, and power consumption.

However, straightforward implementation of the aforementioned nonlinear activation functions in hardware is very expensive because most of these equations require exponentiation and division [6]. Most of accelerators do not implement an ISA [7]–[9] but rather create modules individually, therefore preventing designers from amortizing the costs of physical activation functions. Thus, besides pushing for the efficient execution of the matrix multiplication operations, special attention should also be paid to the other components of the ANN acceleration hardware. This holds true for the activation function. Each neuron in the hidden and output layers needs an activation function. Therefore, small implementation inefficiencies in the activation function can quickly add up. In fact, to achieve a significant speedup, hardware accelerators possess thousands or more processing elements (PEs). Hence, the number of hardware activation function components can be significant, and efforts to optimize activation function circuits could dramatically decrease ANN area and power requirements [10]. For example, if the *tanh* function is implemented using a 10-bit output and 1000 data points, the storage of the function values will require a 10Kb memory structure. Having hundreds of these modules in a design would require multiple megabits of storage. Indeed, in [11], the authors compare 8-bit neurons *ReLU* and *tanh/sigmoid* activation functions. They show that replacing the *ReLU* with *tanh* increases the neuron area by 20% and neuron energy by 36%.

In general, nonlinear functions like *tanh* cannot be effectively approximated using only combinational logic. However, deep neural networks can tolerate low precision operations, therefore lending themselves to such approximations. Using purely combinational logic has the benefits of providing low latency with small area overhead compared conventional ROM-based approaches. We illustrate this point using the *tanh* and *SELU* functions. Their implementations are generalized and open-sourced.

In this work, we explore the design space trade-offs of neural network activation function circuits. In particular, we focus on the efficient implementation of activation functions using purely combinational logic for higher clocking speed and smaller area overhead. The rest of this paper is organized as follows: previous works are introduced in Section II, in Section III and IV we present a detailed implementation of the *SELU* and *tanh* functions, Section V summarizes the experimental results and Section VI concludes this paper.

## II. Related work

Various approaches have been proposed for implementing activation functions in hardware. Generally, these methods fall into two categories: piecewise approximations and look-up table (LUT) based approaches [12]. In this work, we consider the six most commonly used approaches to make the review concise. On the whole, high-fidelity approximations tend to use more resources and have higher latencies, while low-fidelity implementations incur approximation losses but are faster and require fewer hardware resources. In Figure 1, we plot the approximation of the $e^x$ curve with methods 1, 2, 4 and 5. Method 3 (CORDIC algorithm [13]) and method 6 (Optimized

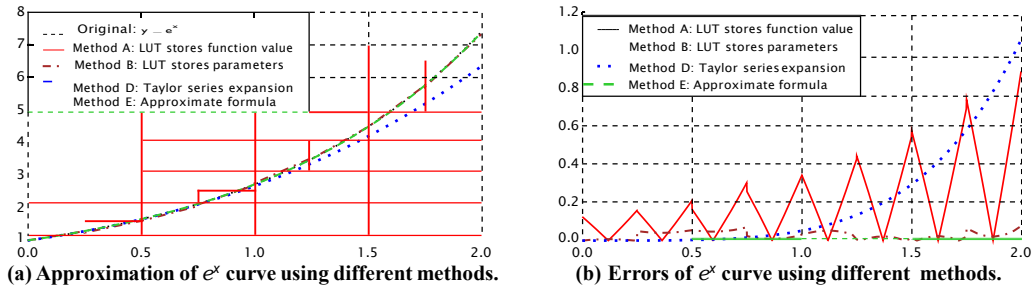

(a) Approximation of $e^x$ curve using different methods.    (b) Errors of $e^x$ curve using different methods.

Fig. 1. **Approximation of $e^x$ curve using different methods.** Method E has a low approximation error, causing the respective curves to be overlapped.

LUT-based method) are omitted as they require too much resource to be directly implemented in hardware.

*A. Storing function values in LUTs*

Look-up tables (LUT) are the most commonly used method to implement activation functions in hardware. The function values are divided into equal sub-ranges and each sub-range is approximated by a value stored in a LUT. For LUT implementations, raising precision requires increasing the sampling rate, adding more storage and increasing latency.

*B. Storing parameters in LUTs*

Instead of storing the function values directly, this method keeps the function slope and the function intercept in the LUT. The function value can then be calculated using the following formula, where $k$ is the slope and $b$ is the function intercept. This approach is a general form of storing function values in LUTs with $k = 0$ (cf. II-A).

$$y = kx + b \quad (1)$$

This method leads to a small improvement on the accuracy, but it also has to store more data and uses an adder and a multiplier to calculate the function values.

*C. CORDIC algorithm*

The third method is the CORDIC algorithm. It uses shift, addition and subtraction operations approximate the nonlinear activation function. The CORDIC algorithm requires less area than storing the parameters in LUTs, but more clock cycles and hardware modules are required to compute the activation function. While the algorithm achieves higher approximation accuracy, its increase in latency may not be suitable for deployment in low-latency edge devices.

*D. Taylor series expansion*

The Taylor series expansion can be used to approximate a nonlinear activation function to any precision. The expansion formula is of the form:

$$f(x) = \sum_{n=0} \frac{f^{(n)}(x_0)}{n!}(x-x_0)^n \quad (2)$$

This method does require multipliers and several clock cycles to perform the calculation.

*E. Approximation formula*

The method introduced in [14] uses the following formula to approximate the exponential function:

$$e^x \approx Ex(x) \approx 2^{1.44x} \quad (3)$$

Based on this formulation, one can calculate the *sigmoid* function as:

$$Sigmoid(x) \approx \frac{1}{1 + 2^{-1.44x}} \approx \frac{1}{1 + 2^{-1.5x}} \quad (4)$$

whereas the *tanh* function can be calculated as:

$$tanh(x) = 1 - 2\,Sigmoid(-2x) \quad (5)$$

The method requires four cycles to approximate the *sigmoid* function. The authors designed a structure to calculate the expression $2^{-1.5x}$, which takes two cycles. An add and a division operations are also performed and take one cycle each. For the *tanh* function approximate an additional clock cycle is required. The implementation of this approximation formula uses fewer resources than the CORDIC approach, but its latency is still high.

*F. Optimized LUT-based method*

This approach is an optimized LUT-based method combined with a Taylor series expansion. The equation is expanded up to the fifth-order:

$$tanh(x) \approx x - \frac{x^3}{3} + \frac{2x^5}{15} \quad (6)$$

When $\frac{x^3}{3} - \frac{2x^5}{15} \leq 0.02$, one can use the approximation $tanh(x) \approx x$. By solving the inequality, one gets $x \geq 0.39$, and $tanh(2.90) \approx 1$. Only the values in the [0.39, 2.90] range need to be stored.

In all, LUT based methods need storage/memory and an extra pipeline stage for the memory access. All these methods, except the one that stores function values in LUTs, either require relatively complex calculations/logic or several clock cycles to minimize the approximation error. According to our experiments in Section V, ANNs are generally insensitive to activation function precision. This is a key insight that allows us to simplify the approximation method without sacrificing the system accuracy. In the following sections, we analyze the activation functions and present our proposed combinational circuit based implementation method.

## III. ACTIVATION FUNCTIONS EXPLORATION

In this section, we discuss the nonlinear activation functions realized using our proposed design approach. We define the *sigmoid* and *tanh* function as:

$$sigmoid(x) = \frac{1}{1+e^{-x}}, \quad tanh(x) = \frac{e^x - e^{-x}}{e^x + e^{-x}} \quad (7)$$

Compared to *sigmoid* function, the *tanh* function passes through zero and can be approximated as $y = x$ around zero. As a result, when the absolute value of the input is small enough, one can perform the matrix operation directly, therefore, the training process is relatively easy. In principle, *sigmoid* and *tanh* have similar expressive ability, but in practice, *sigmoid* is equivalent to an activation function with a bias. It still needs the real bias term to offset its influence, which can affect the optimization. Therefore, the *tanh* function is used more often. Furthermore, it converges faster than the *sigmoid* function.

The *tanh* function has been shown experimentally to outperform the *sigmoid* function. There two reasons for this: the output of the *tanh* function is normalized around 0, producing both positive and negative outputs. The *sigmoid* is not, introducing a systematic bias. Second, when the output of the neuron restricted to [ 1, 1], the activation is more likely to be close to 0, so the neurons are generally less saturated with *tanh* than with *sigmoid*, allowing gradients to better propagate and speeding up learning [15].

In [5], the authors introduced the *SELU* function and analytically proved that neuron activations converge towards zero mean and unit variance. This allows networks with *SELU* activations to train deeper models, speed up learning, and use stronger regularizers without sacrificing accuracy. This is the main motivation behind focusing our work on the efficient implementation of exponential functions.

We define ELU and SELU as:

$$ELU(x) = \begin{cases} x & x > 0 \\ ae^x - a & \text{else} \end{cases} \quad (8)$$

$$SELU(x) = \lambda ELU(x). \quad (9)$$

Here $\lambda = 1.0507$ and $a = 1.6733$.

### IV. ACTIVATION FUNCTION IMPLEMENTATION

#### A. Implementation of the tanh function

In this section, we introduce our method for implementing the *tanh* activation function using exclusively combinational circuits. We consider only the intervals where the function changes significantly.

*1) Properties of the tanh function:* Tanh is an odd function, meaning that it is symmetric with respect to $0$. In order to approximate it, we only need to observe the positive half of the function. As it converges to $1$, we approximate its value in the range $[0, 2]$ for the targeted precision in this work.

We divide the activation function range into $2^k$ segments evenly with the step $\frac{1}{2^k}$. The approximation error depends on the number $k$, which controls the sampling density. The larger the $k$ is, the lower the approximation errors are, but more complex the implementation. During training, the exact *tanh* function is used to calculate the gradients, since the approximate function is non-differentiable. The approximated function is used for the forward pass.

*2) Encoding the value of the activation function:* After selecting a sampling rate, we choose the output value's integer and fractional parts bit-width. The integer part is either $0$ or $1$. For the illustrative case, in order to simplify the complexity of the combinational logic, we choose $7$-bits to encode the output value of the activation function: $1$-bit for the integer part and $6$-bits for the fractional part.

*3) Generating the Karnaugh map for the tanh function:* Boolean functions can be expressed in their canonical form: by listing the input values on the left side of the truth table and the output values on the right side, we get a Karnaugh map. Figure 2 shows the Karnaugh map for one of the *tanh* activation bits. By analyzing the map, one can derive the needed circuit for implementing the bit. We repeat this procedure for all the bits of the *tanh* activation function.

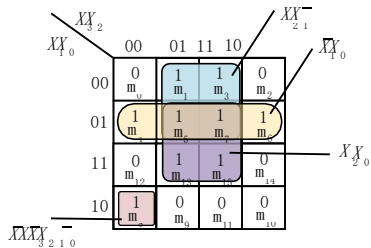

Fig. 2. Karnaugh map of one of the output bits of the *tanh* activation function ($Y_1$) with a $4$-bits input and a $7$-bits output (*tanh 7_4*).

A direct implementation will have a circuit for every cell with the value 1, and a multiple-input OR gate choosing one of these circuit outputs. We can simplify the logic expression from the Karnaugh map by combining some of the adjacent 1's in the table cells. This

### TABLE I
### TANH 7_4 IMPLEMENTATION

| Input | $X_3 X_2 X_1 X_0$ | | |
|---|---|---|---|
| **AND plane** | | | |
| $p_1$ | $\{X_3\}$ | $p_2$ | $\{X_2, X_0\}$ |
| $p_3$ | $\{\overline{X_2}, X_1\}$ | $p_4$ | $\{X_2, X_1, X_0\}$ |
| $p_5$ | $\{X_2, X_1, X_0\}$ | $p_6$ | $\{\overline{X_3}, X_2\}$ |
| $p_7$ | $\{X_3, \overline{X_1}, X_0\}$ | $p_8$ | $\{\overline{X_3}, \overline{X_1}, X_0\}$ |
| $p_9$ | $\{\overline{X_3}, \overline{X_2}, \overline{X_1}, X_0\}$ | $p_{10}$ | $\{\overline{X_3}, X_2, X_0\}$ |
| $p_{11}$ | $\{X_3, X_2, X_1\}$ | $p_{12}$ | $\{\overline{X_3}, \overline{X_1}, X_0\}$ |
| $p_{13}$ | $\{X_2, \overline{X_1}, X_0\}$ | $p_{14}$ | $\{X_3, X_2, \overline{X_1}, X_0\}$ |
| $p_{15}$ | $\{\overline{X_1}, X_0\}$ | $p_{16}$ | $\{X_2, \overline{X_1}\}$ |
| $p_{17}$ | $\{X_3, X_2, X_1, X_0\}$ | $p_{18}$ | $\{X_3, X_1, X_0\}$ |
| $p_{19}$ | $\{X_3, X_2, X_1\}$ | | |
| **Output** | $Y_6 Y_5 Y_4 Y_3 Y_2 Y_1 Y_0$ | | |
| **OR plane** | | | |
| $Y_6$ | $\{p_3, p_2, p_1\}$ | | |
| $Y_5$ | $\{p_5, p_4, p_1\}$ | | |
| $Y_4$ | $\{p_8, p_7, p_6, p_3, p_2\}$ | | |
| $Y_3$ | $\{p_{14}, p_{13}, p_{12}, p_{11}, p_{10}, p_9, p_7\}$ | | |
| $Y_2$ | $\{p_{15}, p_{11}, p_{10}, p_5\}$ | | |
| $Y_1$ | $\{p_{17}, p_{16}, p_{15}, p_2\}$ | | |
| $Y_0$ | $\{p_{19}, p_{18}, p_{11}, p_8, p_4\}$ | | |

reduces the number of individual circuits. As an illustration, here is the expression of one bit of the output value:

$$Y_1 = X_2 \overline{X_1} + X_2 X_0 + \overline{X_3 X_2 X_1 X_0} + \overline{X_1 X_0} \quad (10)$$

$X_i$ refers to the i+1-th bit of the input value, and the $Y_1$ refers to the second bit of the output value.

*4) Combinational logic for the tanh function:* Finally, we can implement the logic expression using an RTL language to get the logical circuits. As for the negative part of the function, since the *tanh* is an odd function, we can deliver the sign bits to the output directly. If we use $g(x)$ to represent the ladder function between $[-2, 2]$, the approximated activation function *tanh* can be written as follows:

$$tanh(x) = \begin{cases} 1 & x \geq 2 \\ g(x) & 2 \geq x > -2 \\ -1 & -2 \geq x \end{cases} \quad (11)$$

*5) Simulation and validation:* Once we have the RTL module, we need to simulate it to check the logic expression and make sure it approximates the desired function. Next, we analyze the time delay of the combinational circuit and check whether the activation function lies on the critical path of the design. After functional and timing testing, if there exist any race conditions or hazards, we change the Karnaugh map to remove them.

After simplifying the logic expression, we obtain the final expressions of the *tanh* function as illustrated in Table I.

#### B. Implementation of the SELU activation function

We demonstrate in this section the implementation of the *SELU* function using only combinational circuits.

*1) Properties of the SELU activation function:* From the formula 8, the positive part of the *SELU* function is linear, so we only need to approximate the negative part. Considering $e^{-3.875} \approx 0.0208 \approx 0$, if the input value is less than $-3.875$, the output value is $-a$, $a$ being a static predetermined parameter. We then divide the interval $[-3.875, 0]$ into $k$ segments evenly.

*2) Encoding the value of the activation function:* We encode the input value with $5$ bits. To maintain precision, we encode the output value into $8$-bits, $1$ bit for the integer part and $7$-bits for the fractional part. In this way, it can be represented as *tanh 8_5*.

*3) Generating the Karnaugh map for the SELU function:* We can construct the truth table and Karnaugh map in a similar fashion as described in Section IV.A.3). From the Karnaugh map, we draw the Karnaugh circle to get the simplest logical expression without race condition and hazards. In total, we arrive at $31$ logical expressions. Here we show an illustration using one bit of the output value:

$$Y_3 = X_4 X_3 X_2 + X_4 X_3 X_1 + X_3 X_1 X_0$$
$$+ \overline{X_2 X_1 X_0} + \overline{X_3 X_2 X_1} + \overline{X_4 X_3 X_2 X_1} \quad (12)$$
$$+ \overline{X_4 X_3 X_1 X_0} + \overline{X_4 X_3 X_1 X_0} + \overline{X_4 X_3 X_2 X_1}$$

$X_i$ refers to the $i+1$-th bit of the input value, and $Y_1$ refers to the second bit of the output value.

In Figure 3, each color block refers to a product, and the logic expression is the sum of all the products. The blocks that have the same color refer to the same product.

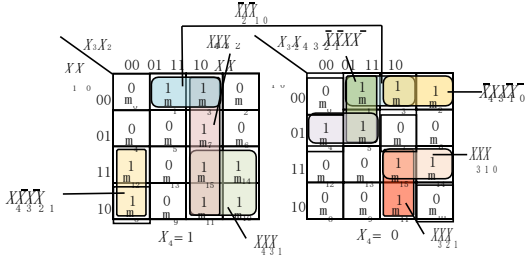

Fig. 3. Karnaugh map of one of the output bits of *SELU* activation function ($y3$) with 5-bits input and 8-bits output(*SELU_8_5*).

*4) Combinational logic of the SELU function:* The approximation of *SELU* using purely combinational logic is shown in Table II. The table shows the final complete logic expressions. We define the SELU function using the formulation shown in equation 13. It is worth noting that we only define it on the $(-3.875, 0)$ interval, as the function is linear for $x \geq 0$ and constant for $x \leq -3.875$.

$$SeLU(x) = \lambda \begin{cases} x & x \geq 0 \\ f(x) & 0 \geq x > -3.875 \\ -a & -3.875 \geq x \end{cases} \quad (13)$$

*5) Simulation and validation:* The purpose of the simulation is the same as in the case of the *tanh* activation function. As more variables may lead to race conditions and hazards more easily, all the possible combinations should be simulated.

More accurate approximations can be achieved by increasing the number of bits for inputs and outputs. Increasing the number of bits in the input helps break the function into more linear segments. Whereas, a larger number of bits in the output representation boosts its precision. In all, using higher bit-widths improves the approximation accuracy but also leads to more complex circuits.

## V. EXPERIMENT

Look-up table based designs are the most common implementation of activation functions. Therefore, in our comparative study, for the baseline designs, we implement the *tanh* and *SELU* functions using look-up tables. The function values storage based implementation is denoted as (ROM_y) and the parameters storage based one is (ROM_k_b). Their construction uses LUTs and follows the procedures described in Section II. Their comparison with the proposed combinational circuit based approach is done in terms of approximation error, power, area, and network accuracy.

The evaluation is conducted in a two-step, software-hardware approach. First, we evaluate the approximation method in software using PyTorch to verify the neural network accuracy. It is worth noting that the procedure may run multiple iterations to find out an appropriate bit-width. Second, for a selected bit-width, the full neural network is implemented in hardware. We then perform circuit-level analysis on the RTL code and deploy it on the FPGA board for further system-level validation.

### A. Approximation Error

The average errors for the three different methods – the proposed combinational circuit based approach, ROM_y and ROM_k_b – are shown in Table III. The errors for the *tanh* and *SELU* functions are bounded to the ranges $-2 < x < 2$ and $-3.875 < x < 0$, respectively. The average error is calculated using the following formula:

$$AverageError = \frac{\sum |P-A|}{N} \times 100\% \quad (14)$$

TABLE II
SELU 8 5 IMPLEMENTATION

| Input | $X_4 X_3 X_2 X_1 X_0$ | | |
|---|---|---|---|
| **AND plane** | | | |
| $p_1$ | $\{X_3\}$ | $p_2$ | $\{X_4\}$ |
| $p_3$ | $\{\underline{X}_4, \underline{X}_3\}$ | $p_4$ | $\{X_2, X_1, \underline{X}_0\}$ |
| $p_5$ | $\{\underline{X}_3, X_2, X_1\}$ | $p_6$ | $\{\underline{X}_4, X_1, \underline{X}_0\}$ |
| $p_7$ | $\{X_4, X_3, X_1\}$ | $p_8$ | $\{X_4, X_3, X_2\}$ |
| $p_9$ | $\{X_4, X_2, X_0\}$ | $p_{10}$ | $\{X_4, X_2, X_1\}$ |
| $p_{11}$ | $\{X_4, \underline{X}_3, \underline{X}_2\}$ | $p_{12}$ | $\{X_4, X_3, X_1\}$ |
| $p_{13}$ | $\{X_2, X_1, \underline{X}_0\}$ | $p_{14}$ | $\{\underline{X}_3, X_2, X_1\}$ |
| $p_{15}$ | $\{X_3, \underline{X}_1, \underline{X}_0\}$ | $p_{16}$ | $\{\underline{X}_4, \underline{X}_3, X_1\}$ |
| $p_{17}$ | $\{\underline{X}_3, X_2, \underline{X}_1\}$ | $p_{18}$ | $\{\underline{X}_3, X_2, \underline{X}_1, X_0\}$ |
| $p_{19}$ | $\{X_3, X_2, X_0\}$ | $p_{20}$ | $\{X_4, X_2, X_0\}$ |
| $p_{21}$ | $\{X_4, X_3, \underline{X}_2, X_0\}$ | $p_{22}$ | $\{X_4, X_2, X_1, X_0\}$ |
| $p_{23}$ | $\{X_4, \underline{X}_3, \underline{X}_1, X_0\}$ | $p_{24}$ | $\{\underline{X}_4, \underline{X}_3, \underline{X}_2, X_1\}$ |
| $p_{25}$ | $\{X_4, \underline{X}_3, \underline{X}_2, X_1\}$ | $p_{26}$ | $\{\underline{X}_4, \underline{X}_3, X_1, \underline{X}_0\}$ |
| $p_{27}$ | $\{X_4, \underline{X}_3, \underline{X}_1, X_0\}$ | $p_{28}$ | $\{X_4, \underline{X}_3, X_2, X_1\}$ |
| $p_{29}$ | $\{X_4, X_3, X_2, X_0\}$ | $p_{30}$ | $\{\underline{X}_4, \underline{X}_3, X_1, \underline{X}_0\}$ |
| $p_{31}$ | $\{X_4, X_2, X_1, X_0\}$ | $p_{32}$ | $\{\underline{X}_4, \underline{X}_3, \underline{X}_2, \underline{X}_0\}$ |
| $p_{33}$ | $\{X_3, \underline{X}_2, X_1, \underline{X}_0\}$ | $p_{34}$ | $\{X_4, \underline{X}_3, \underline{X}_2, X_1\}$ |
| $p_{35}$ | $\{X_4, X_3, X_1, \underline{X}_0\}$ | $p_{36}$ | $\{\underline{X}_4, X_2, \underline{X}_1, \underline{X}_0\}$ |
| $p_{37}$ | $\{X_4, \underline{X}_3, \underline{X}_2, X_1\}$ | $p_{38}$ | $\{\underline{X}_4, \underline{X}_3, X_2, X_1\}$ |
| $p_{39}$ | $\{X_4, \underline{X}_3, X_2, X_0\}$ | $p_{40}$ | $\{\underline{X}_4, \underline{X}_3, X_1, \underline{X}_0\}$ |
| $p_{41}$ | $\{X_4, X_2, \underline{X}_1, X_0\}$ | $p_{42}$ | $\{X_4, \underline{X}_3, X_2, X_1\}$ |
| $p_{43}$ | $\{\underline{X}_3, \underline{X}_2, \underline{X}_1, X_0\}$ | $p_{44}$ | $\{X_3, X_2, \underline{X}_1, \underline{X}_0\}$ |
| $p_{45}$ | $\{X_3, X_2, X_1, X_0\}$ | $p_{46}$ | $\{X_4, \underline{X}_3, \underline{X}_2, X_1\}$ |
| $p_{47}$ | $\{\underline{X}_4, \underline{X}_3, X_1, X_0\}$ | $p_{48}$ | $\{X_4, \underline{X}_3, X_2, \underline{X}_0\}$ |
| $p_{49}$ | $\{\underline{X}_3, \underline{X}_2, X_1, \underline{X}_0\}$ | $p_{50}$ | $\{\underline{X}_3, \underline{X}_2, \underline{X}_1, \underline{X}_0\}$ |
| $p_{51}$ | $\{\underline{X}_4, \underline{X}_3, \underline{X}_2, \underline{X}_1, X_0\}$ | $p_{52}$ | $\{\underline{X}_4, \underline{X}_3, X_2, \underline{X}_1, \underline{X}_0\}$ |
| $p_{53}$ | $\{\underline{X}_4, \underline{X}_3, X_2, \underline{X}_1, X_0\}$ | $p_{54}$ | $\{\underline{X}_4, \underline{X}_3, X_2, \underline{X}_1, \underline{X}_0\}$ |
| $p_{55}$ | $\{X_4, X_3, X_2, X_1, X_0\}$ | $p_{56}$ | $\{X_4, X_3, X_2, X_1, \underline{X}_0\}$ |
| **Output** | $Y_7 Y_6 Y_5 Y_4 Y_3 Y_2 Y_1 Y_0$ | | |
| **OR plane** | | | |
| $Y_7$ | $\{p_4, p_2, p_1\}$ | | |
| $Y_6$ | $\{p_{19}, p_{18}, p_5, p_2\}$ | | |
| $Y_5$ | $\{p_{20}, p_8, p_7, p_6\}$ | | |
| $Y_4$ | $\{p_{53}, p_{52}, p_{51}, p_{24}, p_{23}, p_{22}, p_{21}, p_{10}, p_9, p_3\}$ | | |
| $Y_3$ | $\{p_{28}, p_{27}, p_{26}, p_{25}, p_{15}, p_{14}, p_{13}, p_{12}, p_{11}\}$ | | |
| $Y_2$ | $\{p_{55}, p_{54}, p_{34}, p_{32}, p_{31}, p_{30}, p_{29}, p_{18}, p_{17}, p_{16}\}$ | | |
| $Y_1$ | $\{p_{45}, p_{44}, p_{43}, p_{42}, p_{41}, p_{40}, p_{39}, p_{38}, p_{37}, p_{36}, p_{35}, p_{34}, p_{27}, p_{23}, p_{22}, p_{19}\}$ | | |
| $Y_0$ | $\{p_{56}, p_{50}, p_{49}, p_{48}, p_{47}, p_{47}, p_{46}, p_{44}\}$ | | |

TABLE III
COMPARISON OF AE(AVERAGE ERROR) AND AREA

| Active Function | Tanh_7_4 | | SELU_8_5 | |
|---|---|---|---|---|
| Index | AE | Area($\mu m^2$) | AE | Area($\mu m^2$) |
| Our method | 4.19% | 97.65 | 2.22% | 137.59 |
| ROM_y[1] | 4.19% | 306.12 | 2.22% | 612.24 |
| ROM_k_b | 0.52% | 751.44 | 0.17% | 1162.93 |

P is the function value, A is the approximate value, and N represents the number of sample points. Since piecewise linear approximation is used in our proposed method, the average error is the same as in the function values storage approach (ROM_y) and larger than when the parameters are stored (ROM_k_b). The parameters storage approach does use more resources for this slight accuracy improvement (cf. V-B).

### B. Resources Analysis

To get more accurate results, we synthesized the different designs using a 28nm SVT library. The absolute time delays for the *tanh* and *SELU* functions are 0.1947ns and 0.221ns. This means that the combinational logic can operate at the maximum frequencies of 5.14GHz (*tanh*) and 4.52GHz (*SELU*). The area overheads of the *tanh* and *SELU* implementations are shown in the Table III.

The proposed combinational circuit based method implementation of the *tanh* function saves 68.1% and 87.0% in area compared to function values storage approach (ROM_y) and the parameters storage method (ROM_k_b), respectively. For the *SELU* function implementation, the area savings are 77.53% and 88.17% over ROM_y and ROM_k_b, respectively. The three methods are deployed on the Xilinx VC7V2000T FPGA board. The power consumption results are reported in Figure 4.

---

[1]4-bits inputs and 7-bits outputs (1-bit for the integer) LUT can be implemented with $16 \times 6$bits ROM, which is estimated to take half of the area of a $32 \times 6$bits ROM. The ROM in Memory Compiler has at least 32 entries. 5-bits inputs and 8-bits outputs (1-bit for the integer) LUT can be implemented with $32 \times 8$bits ROM.

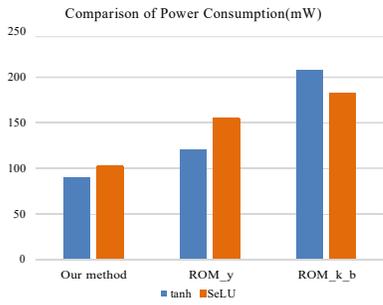

Fig. 4. Power consumption of the three methods. The energy consumption of the proposed method is lower compared to the alternative LUT-based approaches.

One clock cycle is needed to get the function value from the LUT for both the ROM_y and ROM_k_b approaches. For the ROM_k_b, two clock cycles are needed for the linear function computation. On the other hand, the proposed method is purely combinational.

*C. Inference Accuracy*

In this section, we focus on the network accuracy. We use Pytorch with bit-wise operations to approximate the activation functions. We train the neural networks using the original, full precision activation function implementations. Then in the validation phase we replace the activation functions with their approximation function circuits. We make no attempt to retrain the network after changing the activation function. Since such an attempt may remove accuracy losses incurred by quantizing the activation functions.

TABLE IV
PERFORMANCE OF OUR METHOD ON ANN COMPARED WITH ORIGINAL DESIGN

|  | MNIST | CIFAR-10 | ImageNet(Top1/Top5) |
|---|---|---|---|
| Tanh(Original) | 96.15% | 87.17% | 42.39%/67.61% |
| Tanh_5_4 | −0.2% | −5.07% | −8.16%/−8.94% |
| Tanh_7_4 | −0.05% | −1.96% | −7.83%/−8.42% |
| Tanh_7_6 | +0.04% | −0.29% | −7.23%/−8.0% |
| SeLU(Original) | 97.67% | 86.79% | 39.260%/63.342% |
| SeLU_5_4 | +0.04% | −4.15% | −0.122%/+0.458% |
| SeLU_7_4 | +0.01% | −4.47% | −0.004%/+0.866% |
| SeLU_8_5 | +0.37% | −0.69% | +0.368%/+1.278% |

We test the proposed method with LeNet on the MNIST dataset, VGG-16 on the CIFAR-10, and AlexNet on the IMAGENET dataset. The experimental results show an accuracy loss of 0.05% and an increase of 0.37% compared to the original network on MNIST using *tanh_7_4* and *SELU_8_5*, respectively. In case of the CIFAR-10 experiments, we get an accuracy loss of 1.96% and 0.69% for *tanh_7_4* and *SELU_8_5*. For the experiments on the IMAGENET, the accuracy losses are 7.83% on top-1 and 8.42% on top-5 under *tanh_7_4*, while there are gains of 0.368% on top-1 and 1.278% on top-5 for the *SELU_8_5*. The results of the comparative study of the exact implementation and the proposed approximation method are summarized in Table IV. When the quantization method is applied, the network inference accuracy increases. The overall effect of the quantization precision on the inference accuracy follows the pattern observed in other studies [16].

VI. CONCLUSION

In this work, we propose an efficient approximation scheme for activation functions using purely combinational logic, which takes only one clock cycle. We should its implementation and performance on two widely used activation functions, i.e., *tanh* and *SELU*. We conduct a comparative study of the proposed method with other widely used methods, i.e., storage based approaches. Based on the average approximation errors, our method has the best performance to circuit complexity ratio. Activation quantization bears little effect on network accuracy. The hardware implementation of the proposed activation functions is realized using the 28nm SVT library to further validate the efficiency of the proposed approach in terms of area and timing delay. Area reductions of 68.1% and 87.0% for the *tanh* function, and 77.53% and 88.17% for the *SELU* function are recorded when compared with the two baseline LUT-based activation function implementations (ROM y and ROM k b).

VII. ACKNOWLEDGEMENTS

This work was supported in part by the National Science and Technology Major Project of China No.2018ZX01028-101-001, National Key Research and Development Plan No.2016YFB0200202 and National Natural Science Foundation of China No.61773307.